\pdfoutput=1

\documentclass[11pt]{article}
\usepackage[]{ACL2023}
\usepackage{colortbl}
\usepackage{accents}
\usepackage{times}
\usepackage{latexsym}

\usepackage[T1]{fontenc}

\usepackage[utf8]{inputenc}

\DeclareUnicodeCharacter{1E03}{\d{c}}   
\DeclareUnicodeCharacter{1E05}{\d{m}}   
\DeclareUnicodeCharacter{0324}{\d{}}     
\DeclareUnicodeCharacter{0964}{\textbar}

\usepackage{microtype}

\usepackage{inconsolata}

\usepackage{multirow}
\usepackage{multicol}
\usepackage{adjustbox}
\usepackage{natbib}
\usepackage{enumitem}
\usepackage{xcolor}
\usepackage{svg}
\usepackage{siunitx}
\usepackage{graphicx}

\definecolor{orange}{rgb}{1,0.5,0}
\definecolor{mdgreen}{rgb}{0.05,0.6,0.05}
\definecolor{cblue}{rgb}{0.1,0.8,1.0}
\definecolor{dkblue}{rgb}{0,0,0.5}
\definecolor{dkgray}{rgb}{0.3,0.3,0.3}
\definecolor{slate}{rgb}{0.25,0.25,0.4}
\definecolor{gray}{rgb}{0.5,0.5,0.5}
\definecolor{ltgray}{rgb}{0.7,0.7,0.7}
\definecolor{purple}{rgb}{0.7,0,1.0}
\definecolor{lavender}{rgb}{0.65,0.55,1.0}

\title{Mukhyansh: A Headline Generation Dataset for Indic Languages}

\author{Lokesh Madasu\thanks{* Authors contributed equally} ,
  Gopichand Kanumolu\footnotemark[1] ,
  Nirmal Surange\footnotemark[1] ,
  Manish Shrivastava \\
   Language Technologies Research Center, KCIS, IIIT Hyderabad, India.\\
\texttt{\{lokesh.madasu, gopichand.kanumolu, nirmal.surange\}@research.iiit.ac.in} \\
\texttt{m.shrivastava@iiit.ac.in}
}

\begin{document}

\maketitle

\begin{abstract}

The task of headline generation within the realm of Natural Language Processing (NLP) holds immense significance, as it strives to distill the true essence of textual content into concise and attention-grabbing summaries. While noteworthy progress has been made in headline generation for widely spoken languages like English, there persist numerous challenges when it comes to generating headlines in low-resource languages, such as the rich and diverse Indian languages. A prominent obstacle that specifically hinders headline generation in Indian languages is the scarcity of high-quality annotated data. To address this crucial gap, we proudly present Mukhyansh, an extensive multilingual dataset, tailored for Indian language headline generation. Comprising an impressive collection of over 3.39 million article-headline pairs, Mukhyansh spans across eight prominent Indian languages, namely Telugu, Tamil, Kannada, Malayalam, Hindi, Bengali, Marathi, and Gujarati. 
We present a comprehensive evaluation of several state-of-the-art baseline models. Additionally, through an empirical analysis of existing works, we demonstrate that Mukhyansh outperforms all other models, achieving an impressive average ROUGE-L score of 31.43 across all 8 languages.

\end{abstract}

\section{Introduction}

Headline generation plays a crucial role in summarizing news articles and capturing readers' attention. 
The task of headline generation involves automatically generating informative and captivating headlines that accurately capture the essence of the underlying text. 
Headline generation is challenging due to two major factors: firstly, headlines must accurately represent the content of the text while being concise. This requires a fine balance between capturing the key information and maintaining brevity. Secondly, headlines often need to be attention-grabbing, compelling readers to click and read further. This necessitates the use of persuasive language, creativity, and an understanding of rhetorical devices.

In recent years, the NLP community has achieved remarkable strides in the development of headline-generation models. However, the focus has primarily been on English and other widely spoken languages, inadvertently leaving a significant void in the realm of headline generation for Indian languages. While datasets like Gigaword \citep{12.EnglishGigaword, napoles-etal-2012-annotated} have emerged as prominent resources, comprising an impressive collection of over 4 million news article-headline pairs, it is crucial to acknowledge that they are limited to English and fail to capture the intricacies and linguistic nuances of Indian languages.

India, with its rich linguistic diversity, boasts a staggering array of over 22 officially recognized languages, each with its own distinct grammar, syntax, and vocabulary. Addressing the challenge of headline generation in Indian languages necessitates a deep understanding of the specific linguistic and cultural intricacies inherent in each language.

One of the most significant obstacles hindering headline generation in Indian languages is the scarcity of high-quality annotated data. This scarcity severely limits the effectiveness of model training and impedes the performance of supervised learning approaches, which heavily rely on labeled examples.

Fortunately, recent advancements in neural network architectures, such as transformer-based models, have significantly enhanced the performance of headline generation models. These models possess the ability to encode input text and generate headlines by optimizing various objectives, including semantic coherence, informativeness, and readability. While these models have successfully reduced the dependency on labeled data, they still leverage fine-tuning on specialized headline generation datasets to further enhance their performance.

In the context of Bengali language, \citet{22.BengaliHG1,23.BengaliHG2} conducted data collection\footnote{However, the dataset is not made publicly available} from various news websites using web scraping techniques. They proposed an RNN-based encoder-decoder model with an attention mechanism for headline generation. Another notable resource for multilingual abstractive summarization, XL-Sum, was introduced by \citet{27.XL-Sum}. The Indian language section of the XL-Sum dataset consists of 251K article-headline pairs sourced from BBC\footnote{\url{https://www.bbc.com/}}.
To further advance research in Natural Language Generation (NLG) for Indian languages, \citet{13.IndicNLG} proposed the IndicNLG benchmark, encompassing five different NLG tasks, including a headline generation dataset (hereafter referred to as IndicHG dataset). This dataset comprises 1.31 million article-headline pairs across 11 Indian languages. However, our analysis (detailed in Section \ref{sec:IndiCNLG Quality}) reveals serious quality issues, such as data contamination, rendering it unsuitable for training robust models. Despite its claimed size, the dataset's problematic samples significantly reduce its effective size by nearly half.
To summarize our main contributions:
\begin{enumerate}[leftmargin=*]
    \item We present a large, multilingual headline-generation dataset "Mukhyansh", comprising over 3.39 million news article-headline pairs across 8 Indian languages; namely Telugu, Tamil, Kannada, Malayalam, Hindi, Bengali, Marathi, and Gujarati. Our data collection methodology involves developing site-specific crawlers, leveraging a deep understanding of news website structures to ensure the acquisition of high-quality data. 
    \item We employ state-of-the-art baseline models and demonstrate the effectiveness of these models for a diverse range of test sets. 
    \item We provide further evidence to support our argument regarding the necessity of high-quality data by undertaking a comprehensive comparative analysis, specifically contrasting our research with the existing work, particularly IndicHG.

    The dataset and models are available at: \url{https://github.com/ltrc/Mukhyansh}
    
\end{enumerate}

The remaining sections of this paper are structured as follows: Section \ref{Sec:Mukhyansh_intro} provides a comprehensive introduction to Mukhyansh. Section \ref{Sec:baselines} delves into the details of our baseline models. In Section \ref{sec:IndiCNLG Quality}, we meticulously evaluate the existing work, conduct a comparative analysis of each models' performance on diverse datasets, and present our findings. Section \ref{sec:conclusion} concludes with our key contributions, limitations, and future scope.

\section{Mukhyansh \label{Sec:Mukhyansh_intro}}
The data collection process for all eight Indian languages involved web scraping from multiple news websites. However, this task posed challenges due to the diverse and dynamic nature of these websites.

Given that each website has its own unique structure, it was crucial to understand the intricacies of each site to extract data accurately, without any loss of information or introduction of noise. To achieve this, we developed site-specific web scrapers tailored to each website. These scrapers were designed to extract the text of news articles, headlines, and the name of the news subdomain. Care was taken to ensure that both the article and headline elements were non-empty and devoid of any unwanted information such as advertisements, URLs pointing to related articles, or embedded social media content.

To avoid any bias towards a particular news style, data was collected from a diverse range of news websites\footnote{See Appendix \ref{appendix:A} for a detailed list of websites used for scraping}. These websites covered various domains, including state, national, international, entertainment, sports, business, politics, crime, and COVID-19, among others\footnote{Refer to Table \ref{table:Category wise statistics of Mukhyansh} in Appendix for category-wise statistics of the dataset.}.
To ensure the quality of the collected data, additional preprocessing steps were implemented next.

\begin{table*}[h!]
\begin{adjustbox}{width=\textwidth, center}
\begin{tabular}{lrrrr|rrrr}
\hline
\multicolumn{1}{c}{}                  & \multicolumn{4}{c|}{Dravidian language family} & \multicolumn{4}{c}{Indo-Aryan language family} \\ \cline{2-9}
\multicolumn{1}{c}{}                  & te           & ta          & kn          & ml          & hi           & bn           & mr          & gu          \\ \hline
\# Pairs collected           & 1080665      & 378545      & 505641      & 435896      & 729950       & 309008       & 411566      & 338502      \\
\# Duplicates                         & 8024         & 11546       & 64116       & 269         & 32539        & 7055         & 10184       & 35518       \\
\textbf{\# Pairs after deduplication} & 1072641      & 366999      & 441525      & 435627      & 697411       & 301953       & 401382      & 302984      \\ \hline
\# Pairs with prefix                  & 8756         & 1712        & 1983        & 21633       & 2656         & 1302         & 942         & 200         \\
\# Pairs with multiple-articles       & 582          & 0           & 0           & 0           & 0            & 0            & 0           & 0           \\
\# Pairs too short                    & 146181       & 33579       & 101619      & 98921       & 94132        & 19378        & 65998       & 26826       \\
\textbf{\# Pairs after filtering}     & 917122       & 331708      & 337923      & 315072      & 600623       & 281273       & 334442      & 275958      \\ \hline
\# Pairs in train                     & 825372       & 298543      & 304122      & 283555      & 540568       & 253139       & 301001      & 248367      \\
\# Pairs in dev                       & 82571        & 26539       & 27044       & 25190       & 48042        & 22514        & 26751       & 22073       \\
\# Pairs in test                      & 9179         & 6626        & 6757        & 6327        & 12013        & 5620         & 6690        & 5518        \\ \hline    
\end{tabular}
\end{adjustbox}
\caption{Statistics of Mukhyansh Preprocessing.}
\label{table:Mukhyansh_Statistics}
\end{table*}

\subsection{Preprocessing } In the series of essential preprocessing steps, firstly, we eliminate all special symbols, emojis, and punctuation marks from the dataset. Next, we remove any duplicate article-headline pairs from the dataset.
Lead or prefix, wherein the title of an article is derived from the initial sections that typically contain the most crucial information, is a widespread approach adopted by news sites. Although utilizing the lead section can be beneficial for summary generation, it may inadvertently hinder the model's ability to learn and discriminate between different types of information. By relying solely on the lead, the model may overlook relevant details and nuances present in the subsequent sections of the article. Therefore, we eliminate pairs with prefixes from the dataset.
Furthermore, to ensure that only substantial and informative pairs are retained, we apply a minimum-length filter to the dataset. 
This filter helps eliminate article-headline pairs where the article contains fewer than 20 tokens and/or the headline consists of fewer than 3 tokens. Table \ref{table:Mukhyansh_Statistics} provides an overview of the preprocessing statistics for Mukhyansh and the final Train, Dev, and Test splits. 

For the final splits, we allocated 90\% of the data for training purposes, while the remaining data was dedicated to development and testing. To ensure robust performance and prevent any bias towards specific news categories or domains, stratified sampling techniques were employed when creating our data splits. This approach guarantees that articles from all categories are evenly distributed across the training, development, and test sets.
Additional statistical details of the Mukhyansh dataset can be found in Table \ref{tab:mukhyansh_detailed_stats}.

\subsection{Human Evaluation} 
In order to evaluate the quality of the Mukhyansh dataset more comprehensively, a human evaluation was conducted. Due to resource constraints and the expenses associated with annotation, this evaluation was limited to the Telugu language data. A total of 500 article-headline pairs were randomly selected and assigned to native-language annotators. They were provided with a set of guidelines, which were based on those utilized in previous studies such as XL-Sum \citep{27.XL-Sum} and IndicNLG \citep{13.IndicNLG}. The evaluation specifically focused on the following properties:

\begin{itemize}[nosep]
    \item \textbf{Consistent } \textit{True}, If the article and headline are consistent.
    \item \textbf{Inconsistent } \textit{True}, If the headline contains information that is inconsistent with the article.
    \item \textbf{Unfounded} \textit{True}, If the headline contains extra information that cannot be inferred from the article.
\end{itemize}
We assign each article-headline pair to 3 annotators and the final rating for each pair is selected based on majority voting. We found that 96.8\% of the samples were rated \textit{True} for \textit{Consistency}, and the percentage of samples that are rated \textit{Inconsistent}, and \textit{Unfounded} were 0.6\%, and 2.6\% respectively, which supports our claim of a reliable and good-quality dataset.

The inter-annotator agreement was assessed using a variation of Fleiss' Kappa, proposed by \citep{randolph2005free} and it resulted in an encouragingly high score of 0.76, indicating substantial agreement among annotators.

\begin{table*}[h!]
\begin{adjustbox}{width=\textwidth, center}
\begin{tabular}{c||ccc||ccc||ccc||ccc||ccc}
\hline
\multirow{3}{*}{\textbf{L}} & \multicolumn{3}{c||}{\multirow{2}{*}{\textbf{FastText+GRU}}}     & \multicolumn{3}{c||}{\multirow{2}{*}{\textbf{FastText+LSTM}}}    & \multicolumn{3}{c||}{\multirow{2}{*}{\textbf{BPEmb+GRU}}}        & \multicolumn{3}{c||}{\multirow{2}{*}{\textbf{mT5-small}}}        & \multicolumn{3}{c}{\multirow{2}{*}{\textbf{SSIB}}}  \\
                  & \multicolumn{3}{c||}{}                                           & \multicolumn{3}{c||}{}                                           & \multicolumn{3}{c||}{}                                           & \multicolumn{3}{c||}{}                                           & \multicolumn{3}{c}{}                                     \\ \cline{2-16} 
                  & \multicolumn{1}{c|}{\textbf{R-1}}   & \multicolumn{1}{c|}{\textbf{R-2}}   & \textbf{R-L}   & \multicolumn{1}{c|}{\textbf{R-1}}   & \multicolumn{1}{c|}{\textbf{R-2}}   & \textbf{R-L}   & \multicolumn{1}{c|}{\textbf{R-1}}   & \multicolumn{1}{c|}{\textbf{R-2}}   & \textbf{R-L}   & \multicolumn{1}{c|}{\textbf{R-1}}   & \multicolumn{1}{c|}{\textbf{R-2}}   & \textbf{R-L}   & \multicolumn{1}{c|}{\textbf{R-1}} & \multicolumn{1}{c|}{\textbf{R-2}} & \textbf{R-L} \\ \hline
te       & \multicolumn{1}{c|}{32.71} & \multicolumn{1}{c|}{15.00}    & 32.02 & \multicolumn{1}{c|}{33.41} & \multicolumn{1}{c|}{14.93} & 32.70  & \multicolumn{1}{c|}{30.06} & \multicolumn{1}{c|}{14.52} & 29.31 & \multicolumn{1}{c|}{39.34}      & \multicolumn{1}{c|}{21.95} & \textbf{38.35} & \multicolumn{1}{c|}{38.42}    & \multicolumn{1}{c|}{20.85} & 37.33   \\ 
ta       & \multicolumn{1}{c|}{33.52} & \multicolumn{1}{c|}{15.40}  & 32.20  & \multicolumn{1}{c|}{32.64} & \multicolumn{1}{c|}{13.60}  & 31.26 & \multicolumn{1}{c|}{33.28} & \multicolumn{1}{c|}{16.15} & 32.04 & \multicolumn{1}{c|}{43.22}    & \multicolumn{1}{c|}{24.38} & \textbf{41.18} & \multicolumn{1}{c|}{43.47}    & \multicolumn{1}{c|}{24.50} & 41.16   \\ 
kn       & \multicolumn{1}{c|}{26.19} & \multicolumn{1}{c|}{10.53} & 25.25 & \multicolumn{1}{c|}{23.75} & \multicolumn{1}{c|}{\phantom{0}7.94}  & 22.84 & \multicolumn{1}{c|}{24.46} & \multicolumn{1}{c|}{10.68} & 23.60  & \multicolumn{1}{c|}{34.73} & \multicolumn{1}{c|}{17.88} & \textbf{33.34} & \multicolumn{1}{c|}{34.36}    & \multicolumn{1}{c|}{17.06} & 32.59  \\ 
ml       & \multicolumn{1}{c|}{28.86} & \multicolumn{1}{c|}{13.17} & 28.17 & \multicolumn{1}{c|}{24.00}    & \multicolumn{1}{c|}{\phantom{0}8.80}   & 23.44 & \multicolumn{1}{c|}{26.13} & \multicolumn{1}{c|}{13.22} & 25.36 & \multicolumn{1}{c|}{35.50}      & \multicolumn{1}{c|}{20.79} &  \textbf{34.63} & \multicolumn{1}{c|}{33.21}  & \multicolumn{1}{c|}{18.57} & 32.04  \\ 
hi       & \multicolumn{1}{c|}{32.97} & \multicolumn{1}{c|}{14.20}  & 29.50  & \multicolumn{1}{c|}{32.34} & \multicolumn{1}{c|}{11.79} & 28.45 & \multicolumn{1}{c|}{32.24} & \multicolumn{1}{c|}{13.93} & 28.94 & \multicolumn{1}{c|}{38.26}      & \multicolumn{1}{c|}{18.81} &  33.65 & \multicolumn{1}{c|}{41.05}    & \multicolumn{1}{c|}{20.77} & \textbf{36.18}   \\ 
bn       & \multicolumn{1}{c|}{18.55} & \multicolumn{1}{c|}{\phantom{0}6.15}  & 17.47 & \multicolumn{1}{c|}{15.73} & \multicolumn{1}{c|}{\phantom{0}4.00}     & 14.90  & \multicolumn{1}{c|}{10.20}  & \multicolumn{1}{c|}{\phantom{0}2.31}  & \phantom{0}9.84  & \multicolumn{1}{c|}{22.90}      & \multicolumn{1}{c|}{\phantom{0}8.87} &  21.56 & \multicolumn{1}{c|}{23.67}    & \multicolumn{1}{c|}{\phantom{0}8.84} & \textbf{22.04} \\ 
mr       & \multicolumn{1}{c|}{17.26} & \multicolumn{1}{c|}{\phantom{0}5.08}  & 16.83 & \multicolumn{1}{c|}{14.32} & \multicolumn{1}{c|}{\phantom{0}3.11}  & 14.04 & \multicolumn{1}{c|}{17.91} & \multicolumn{1}{c|}{\phantom{0}6.48}  & 17.54 & \multicolumn{1}{c|}{27.25}      & \multicolumn{1}{c|}{12.68}  &  26.41 & \multicolumn{1}{c|}{28.21}    & \multicolumn{1}{c|}{12.95} & \textbf{27.08}     \\ 
gu       & \multicolumn{1}{c|}{15.61} & \multicolumn{1}{c|}{\phantom{0}3.87}  & 14.84 & \multicolumn{1}{c|}{\phantom{0}9.98}  & \multicolumn{1}{c|}{\phantom{0}1.68}  & \phantom{0}9.48  & \multicolumn{1}{c|}{15.68} & \multicolumn{1}{c|}{\phantom{0}4.59}  & 14.94 & \multicolumn{1}{c|}{21.80}      & \multicolumn{1}{c|}{\phantom{0}8.53} & 20.43 & \multicolumn{1}{c|}{24.77}    & \multicolumn{1}{c|}{\phantom{0}9.86} & \textbf{23.05}    \\ \hline
\textbf{Average}  & \multicolumn{1}{c|}{25.71} & \multicolumn{1}{c|}{10.43}  & 24.54 & \multicolumn{1}{c|}{23.27}  & \multicolumn{1}{c|}{\phantom{0}8.23}  & 22.14  & \multicolumn{1}{c|}{23.75} & \multicolumn{1}{c|}{10.24}  & 22.70 & \multicolumn{1}{c|}{32.88}      & \multicolumn{1}{c|}{16.74} & 31.19 & \multicolumn{1}{c|}{33.40}    & \multicolumn{1}{c|}{16.68} & 31.43    \\ \hline
\end{tabular}
\end{adjustbox}
\caption{ROUGE-1,2,L scores of various baseline models of Mukhyansh for each language (L).} 
\label{table:Model Results}
\end{table*}

\section{Baseline Models}
\label{Sec:baselines}
In our research paper, we evaluate the performance of commonly used sequence-to-sequence models as baselines on our dataset. Our implementation includes two categories of models: one based on an RNN encoder-decoder network trained from scratch, and another utilizing fine-tuning with pre-trained transformer encoder-decoder models like mT5 \citep{25.mt5_small} and IndicBART \citep{24.IndicBART}.

For the RNN architecture, we adopt the recurrent neural network proposed by \citet{1.sutskever2014sequence}, with a simple context attention mechanism inspired by  \citet{5.HeadlineGenStanford}, which is a modification of the dot product attention mechanism introduced by \citet{6.ManningAttention}. We explore two variations of this model: one using GRU \citep{3.GRU} in both the encoder and decoder, and the other utilizing LSTM \citep{2.LSTM}.

To tackle the challenge of out-of-vocabulary (OOV) words, particularly prevalent in morphologically rich Indian languages, we employ Byte Pair Encoding (BPE) \citep{10.BPE}. Specifically, we use the GRU architecture\footnote{GRUs use fewer parameters, making them more computationally efficient for our experiments, with limited compute resources.} mentioned earlier and initialize the model with 300d subword embeddings from BPEmb \citep{11.bpemb}.

In addition to the above approaches, we also leverage the benefits of transfer learning in headline generation by utilizing pre-trained sequence-to-sequence models such as mT5 and IndicBART. To implement these models, we utilize the scripts\footnote{ \url{https://github.com/huggingface/transformers/tree/main/examples/pytorch/summarization}} provided by Huggingface \citep{HuggingfaceTransformers}. 

\noindent\textbf{mT5:} mT5 is a multilingual variant of T5 \citep{29.T5} covering 101 languages. For our baseline, we fine-tune the pre-trained mT5-small model on our dataset.

\noindent\textbf{IndicBART:} IndicBART is a multilingual, sequence-to-sequence pre-trained model focusing on 11 Indian languages and English. It is similar to mBART \citep{30.mBART} in terms of architecture and training methodology. Specifically, we use a variant of IndicBART called separate script IndicBART\footnote{\url{https://huggingface.co/ai4bharat/IndicBARTSS}} (hereafter referred to as SSIB) and fine-tune it on our dataset for the task of headline generation.

\begin{table}[]
\resizebox{\columnwidth}{!}{
\begin{tabular}{l|c|c|c|c}
\hline
\textbf{Parameters} & \textbf{\begin{tabular}[c]{@{}c@{}}Seq-Seq\\+\\ FastText\end{tabular}} & \textbf{\begin{tabular}[c]{@{}c@{}}Seq-Seq\\+\\ BPEmb\end{tabular}} & \textbf{mT5-small} & \textbf{SSIB} \\ \hline
Max Source Length  & 200                                                                  & 300                                                               & 1024               & 1024               \\ 
Max Target Length  & 20                                                                   & 30                                                                & 30                 & 30                 \\ 
Vocabulary Size    & 40000                                                                & 40000                                                             & 250112             & 64000              \\ 
Beam Width         & 5                                                                    & 5                                                                 & 4                  & 4                  \\ 
Batch Size         & 16                                                                   & 16                                                                & 16                 & 16                  \\ 
Optimizer          & Adam                                                                 & Adam                                                              & Adam               & Adam               \\ 
Learning rate      & $1e^{-4}$                                                            & $1e^{-4}$                                                         & $5e^{-5}$          & $5e^{-5}$            \\ 
(GPU,CPU)          & (1,10)                                                               & (1,10)                                                            & (4,40)             & (4,40)             \\ \hline
\end{tabular}
}
\caption{Experimental setup of various baseline models.}
\label{table:parameters}
\end{table}

\subsection{Experimental Setup}
The LSTM and GRU models used in this research paper consist of 4 stacked layers, with each LSTM/GRU cell containing 600 hidden activation units. To initialize the word embeddings, we employ the 300d pre-trained FastText embeddings \citep{7.Fasttext} for each language.

During the inference phase, we utilize the beam search strategy with length normalization penalty \citep{26.GoogleNMTBeamSearch}. After conducting experiments with various penalty values, we found that a penalty of 0.1 for Telugu, Tamil, Kannada, and Malayalam, and no length normalization for other languages, yielded superior results. To prevent overfitting, we employ early stopping.

Conversely, due to limited computational resources, for the pre-trained models we fine-tuned them on our data for 10 epochs. The model checkpoint with the highest validation score is selected to generate predictions on the test set.

To assess the models' performance, we utilize the multilingual ROUGE metric \citep{27.XL-Sum}\footnote{\url{https://github.com/csebuetnlp/xl-sum/tree/master/multilingual_rouge_scoring}}. Further details regarding the experimental setup and parameter configurations for all the models can be found in Table \ref{table:parameters}.

\subsection{Results} 
Table \ref{table:Model Results} presents the ROUGE-1, 2, L (R-1, R-2, R-L) scores achieved by different baseline models on Mukhyansh. The best R-L score for each language is highlighted in bold. Notably, the SSIB and mT5-small models outperformed all the sequence-to-sequence models trained from scratch. The superior performance of SSIB and mT5-small can be attributed to their pre-training on a large corpus. 

It is worth mentioning that the GRU variant of the sequence-to-sequence model, utilizing FastText embeddings, yielded satisfactory results with a smaller parameter count (64 Million) compared to SSIB (244 Million) and mT5-small (300 Million).

\section{Existing Dataset Evaluation} \label{sec:IndiCNLG Quality}
Due to the unavailability of publicly accessible data from existing monolingual works, our evaluation is limited to the recent multilingual datasets, namely XL-Sum and IndicHG. While XL-Sum focuses on extreme summarization, it is important to note that the summaries provided may consist of more than one sentence. Additionally, concerns have been raised by \citet{urlana-etal-2022-tesum} regarding the quality of summaries in the Indian language section of XL-Sum. Consequently, our evaluation is primarily centered on the IndicHG dataset\footnote{IndicNLG data for Headline-generation was taken from \url{https://huggingface.co/datasets/ai4bharat/IndicHeadlineGeneration/tree/main/data}}.

To validate the reported results in IndicNLG regarding headline generation, we conduct a series of experiments on the IndicHG dataset, accompanied by comprehensive quantitative and qualitative analyses. As discussed in the subsequent sub-sections, our investigation has uncovered significant quality issues with the HG dataset of IndicNLG.
Despite the valuable contributions of IndicNLG to the field of language generation for various Indic languages, it is imperative to address these issues before deeming the IndicHG dataset suitable for training robust models.

\begin{table}[t]
\resizebox{\columnwidth}{!}{
\begin{tabular}{cccc}
\hline
\multicolumn{4}{c}{\textbf{IndicHG Performance}}                  \\ \hline
\multicolumn{1}{c|}{\textbf{L}}       & \multicolumn{1}{c|}{\textbf{\begin{tabular}[c]{@{}c@{}}Reported\end{tabular}}} & \multicolumn{1}{c|}{\textbf{\begin{tabular}[c]{@{}c@{}}Reproduced\end{tabular}}} & \textbf{\begin{tabular}[c]{@{}c@{}}Unbiased\end{tabular}} \\ \hline
\multicolumn{1}{c|}{te}               & \multicolumn{1}{c|}{41.97}                                                                & \multicolumn{1}{c|}{22.37}                                                                                   & 19.47                                                                                     \\ 
\multicolumn{1}{c|}{ta}               & \multicolumn{1}{c|}{46.52}                                                                & \multicolumn{1}{c|}{32.96}                                                                                   & 33.79                                                                                     \\ 
\multicolumn{1}{c|}{kn}               & \multicolumn{1}{c|}{73.19}                                                                & \multicolumn{1}{c|}{42.79}                                                                                   & 21.64                                                                                     \\ 
\multicolumn{1}{c|}{ml}               & \multicolumn{1}{c|}{60.51}                                                                & \multicolumn{1}{c|}{35.64}                                                                                   & 26.79                                                                                    \\ 
\multicolumn{1}{c|}{hi}               & \multicolumn{1}{c|}{34.49}                                                                & \multicolumn{1}{c|}{24.12}                                                                                   & 22.68                                                                                     \\ 
\multicolumn{1}{c|}{bn}               & \multicolumn{1}{c|}{37.95}                                                                & \multicolumn{1}{c|}{22.54}                                                                                   & 20.28                                                                                     \\ 
\multicolumn{1}{c|}{mr}               & \multicolumn{1}{c|}{40.78}                                                                & \multicolumn{1}{c|}{21.28}                                                                                   & 20.14                                                                                     \\ 
\multicolumn{1}{c|}{gu}               & \multicolumn{1}{c|}{31.80}                                                                 & \multicolumn{1}{c|}{22.68}                                                                                   & 22.61                                                                                     \\ \hline 
\multicolumn{1}{c|}{\textbf{Average}} & \multicolumn{1}{c|}{45.90}                                                                 & \multicolumn{1}{c|}{28.05}                                                                                   & 23.42                                                                                     \\ \hline \hline
\multicolumn{2}{r}{\textbf{Performance drop}}                                                                                   & \multicolumn{1}{c|}{17.85}                                                                                   & 22.48                                                                                     \\ \hline
\end{tabular}
}
\caption{Performance Comparison of various versions of IndicHG: Reported, IndicHG* and IndicHG\_Unbiased.}
\label{table:Comparison_reproducedIndicNLG}
\end{table}

\subsection{Reproducing IndicHG Results}
We initiate our experiments with an attempt to replicate the findings of IndicHG for the eight Indian languages mentioned. Following their paper's methodology and hyper-parameter settings, we meticulously fine-tune the SSIB model, (hereafter, referred to as \textit{IndicHG*}). In order to obtain a more reliable assessment of the model's performance and evaluate the consistency of the results, we conducted the same experiment five times with different initial seeds. Subsequently, we calculate the mean and standard deviation of the ROUGE-L scores\footnote{Due to space constraints, additional details and the corresponding ROUGE-1, ROUGE-2 scores are reported in Appendix \ref{appendix:B}, Table \ref{table:Means_Reproduced_IndicHG_Dataset}} obtained on the test set. Table
\ref{table:Comparison_reproducedIndicNLG} presents these mean ROUGE-L scores alongside their reported\footnote{The reported scores are taken from the monolingual works of IndicHG \cite{13.IndicNLG} paper, as the checkpoint is not made public.} counterparts.

As depicted in the final row of Table  \ref{table:Comparison_reproducedIndicNLG}, there is an average reduction of 17.85 in the ROUGE-L scores across the eight languages. This substantial decrease raises concerns regarding the reproducibility of the original findings and emphasizes the necessity for further investigation.

\begin{table*}[h!]
\begin{adjustbox}{width=\textwidth, center}
\begin{tabular}{cccccccccc|rr}
\hline
\multicolumn{1}{c|}{\multirow{3}{*}{\textbf{L}}}                        & 
        \multicolumn{2}{c||}{\textbf{Train set}}                    & 
        \multicolumn{3}{c||}{\textbf{Development set}}                 & 
        \multicolumn{3}{c||}{\textbf{Test set}}                     & 
        \multicolumn{3}{c}{\textbf{Total}}                      
\\ \cline{2-12} 
\multicolumn{1}{c|}{}                                   & \multicolumn{1}{c|}{\textbf{\# Pairs}} & \multicolumn{1}{c||}{\textbf{\begin{tabular}[c]{@{}c@{}}Duplicates \\ (\%)\end{tabular}}} & \multicolumn{1}{c|}{\textbf{\# Pairs}} & \multicolumn{1}{c|}{\textbf{\begin{tabular}[c]{@{}c@{}}Duplicates \\ (\%)\end{tabular}}} & \multicolumn{1}{c||}{\textbf{\begin{tabular}[c]{@{}c@{}}Train \\ Overlap (\%)\end{tabular}}} & \multicolumn{1}{c|}{\textbf{\# Pairs}} & \multicolumn{1}{c|}{\textbf{\begin{tabular}[c]{@{}c@{}}Duplicates \\ (\%)\end{tabular}}} & \multicolumn{1}{c||}{\textbf{\begin{tabular}[c]{@{}c@{}}Train-Dev \\ Overlap (\%)\end{tabular}}} & \multicolumn{1}{c|}{\textbf{\begin{tabular}[c]{@{}c@{}}\textbf{\# Pairs}\end{tabular}}} & 
\multicolumn{1}{c|}{\textbf{\begin{tabular}[c|]{@{}c@{}}(Duplicates + \\ Overlap) (\%)\end{tabular}}} &
\multicolumn{1}{c}{\textbf{\begin{tabular}[c|]{@{}c@{}} Remaining\end{tabular}}} 
\\ \hline
\multicolumn{1}{c|}{\textbf{te}}                        & \multicolumn{1}{r|}{21352}             & \multicolumn{1}{c||}{\phantom{0}8.77}                                                                & \multicolumn{1}{r|}{2690}              & \multicolumn{1}{c|}{\phantom{0}1.52}                                                                & \multicolumn{1}{c||}{15.61}                                                                      & \multicolumn{1}{r|}{2675}              & \multicolumn{1}{c|}{\phantom{0}1.42}                                                                & \multicolumn{1}{c||}{18.61}                                                                        & \multicolumn{1}{c|}{26717}                                                                      & \multicolumn{1}{c|}{10.38}             & 23945                                                               \\ 
\multicolumn{1}{c|}{\textbf{ta}}                        & \multicolumn{1}{r|}{60650}             & \multicolumn{1}{c||}{51.18}                                                               & \multicolumn{1}{r|}{7616}              & \multicolumn{1}{c|}{50.22}                                                               & \multicolumn{1}{c||}{\phantom{0}3.31}                                                                       & \multicolumn{1}{r|}{7688}              & \multicolumn{1}{c|}{50.20}                                                                & \multicolumn{1}{c||}{\phantom{0}3.62}                                                                        & \multicolumn{1}{c|}{75954}                                                                      & \multicolumn{1}{c|}{51.29}             & 36996                                                               \\ 
\multicolumn{1}{c|}{\textbf{kn}}                        & \multicolumn{1}{r|}{132380}            & \multicolumn{1}{c||}{87.26}                                                               & \multicolumn{1}{r|}{19416}             & \multicolumn{1}{c|}{84.29}                                                               & \multicolumn{1}{c||}{59.18}                                                                      & \multicolumn{1}{r|}{3261}              & \multicolumn{1}{c|}{\phantom{0}6.23}                                                                & \multicolumn{1}{c||}{71.17}                                                                       & \multicolumn{1}{c|}{155057}                                                                     & \multicolumn{1}{c|}{87.51}            & 19364                                                               \\ 
\multicolumn{1}{c|}{\textbf{ml}}                        & \multicolumn{1}{r|}{10358}             & \multicolumn{1}{c||}{22.83}                                                               & \multicolumn{1}{r|}{5388}              & \multicolumn{1}{c|}{76.26}                                                               & \multicolumn{1}{c||}{33.33}                                                                      & \multicolumn{1}{r|}{5220}              & \multicolumn{1}{c|}{76.05}                                                               & \multicolumn{1}{c||}{44.22}                                                                       & \multicolumn{1}{c|}{20966}                                                                      & \multicolumn{1}{c|}{53.78}             & 9690                                                               \\ 
\multicolumn{1}{c|}{\textbf{hi}}                        & \multicolumn{1}{r|}{208091}            & \multicolumn{1}{c||}{\phantom{0}3.19}                                                                & \multicolumn{1}{r|}{44718}             & \multicolumn{1}{c|}{\phantom{0}0.76}                                                                & \multicolumn{1}{c||}{\phantom{0}6.42}                                                                       & \multicolumn{1}{r|}{44475}             & \multicolumn{1}{c|}{\phantom{0}0.72}                                                                & \multicolumn{1}{c||}{\phantom{0}7.83}                                                                        & \multicolumn{1}{c|}{297284}                                                                      & \multicolumn{1}{c|}{\phantom{0}4.59}            & 283646                                                                \\ 
\multicolumn{1}{c|}{\textbf{bn}}                        & \multicolumn{1}{r|}{113424}            & \multicolumn{1}{c||}{69.86}                                                               & \multicolumn{1}{r|}{14739}             & \multicolumn{1}{c|}{68.02}                                                               & \multicolumn{1}{c||}{19.41}                                                                      & \multicolumn{1}{r|}{14568}             & \multicolumn{1}{c|}{67.94}                                                               & \multicolumn{1}{c||}{24.30}                                                                       & \multicolumn{1}{c|}{142731}                                                                      & \multicolumn{1}{c|}{70.65}            & 41896                                                               \\ 
\multicolumn{1}{c|}{\textbf{mr}}                        & \multicolumn{1}{r|}{114000}            & \multicolumn{1}{c||}{69.10}                                                               & \multicolumn{1}{r|}{14250}             & \multicolumn{1}{c|}{66.95}                                                               & \multicolumn{1}{c||}{15.45}                                                                      & \multicolumn{1}{r|}{14340}             & \multicolumn{1}{c|}{67.03}                                                               & \multicolumn{1}{c||}{16.15}                                                                       & \multicolumn{1}{c|}{142590}                                                                      & \multicolumn{1}{c|}{69.73}            & 43157                                                               \\ 
\multicolumn{1}{c|}{\textbf{gu}}                        & \multicolumn{1}{r|}{199972}            & \multicolumn{1}{c||}{75.11}                                                               & \multicolumn{1}{r|}{31270}             & \multicolumn{1}{c|}{80.04}                                                               & \multicolumn{1}{c||}{\phantom{0}0.96}                                                                       & \multicolumn{1}{r|}{31215}             & \multicolumn{1}{c|}{80.02}                                                               & \multicolumn{1}{c||}{\phantom{0}1.28}                                                                        & \multicolumn{1}{c|}{262457}                                                                      & \multicolumn{1}{c|}{76.33}            & 62123                                                               \\ 
\multicolumn{1}{c|}{\textbf{pa}}                        & \multicolumn{1}{r|}{48441}             & \multicolumn{1}{c||}{\phantom{0}0.13}                                                                & \multicolumn{1}{r|}{6108}              & \multicolumn{1}{c|}{\phantom{0}0}                                                                   & \multicolumn{1}{c||}{\phantom{0}0.18}                                                                       & \multicolumn{1}{r|}{6086}              & \multicolumn{1}{c|}{\phantom{0}0}                                                                   & \multicolumn{1}{c||}{\phantom{0}0.35}                                                                        & \multicolumn{1}{c|}{60635}                                                                      & \multicolumn{1}{c|}{\phantom{0}0.16}             &  60540                                                               \\ 
\multicolumn{1}{c|}{\textbf{as}}                        & \multicolumn{1}{r|}{29631}             & \multicolumn{1}{c||}{30.05}                                                               & \multicolumn{1}{r|}{14592}             & \multicolumn{1}{c|}{75.96}                                                               & \multicolumn{1}{c||}{58.77}                                                                      & \multicolumn{1}{r|}{14808}             & \multicolumn{1}{c|}{75.97}                                                               & \multicolumn{1}{c||}{65.91}                                                                       & \multicolumn{1}{c|}{59031}                                                                      & \multicolumn{1}{c|}{60.66}             & 23222                                                               \\ 
\multicolumn{1}{c|}{\textbf{or}}                        & \multicolumn{1}{r|}{58225}             & \multicolumn{1}{c||}{48.77}                                                               & \multicolumn{1}{r|}{7484}              & \multicolumn{1}{c|}{48.97}                                                               & \multicolumn{1}{c||}{\phantom{0}0.16}                                                                       & \multicolumn{1}{r|}{7137}              & \multicolumn{1}{c|}{48.58}                                                               & \multicolumn{1}{c||}{\phantom{0}0.42}                                                                        & \multicolumn{1}{c|}{72846}                                                                      & \multicolumn{1}{c|}{48.79}             &  37305                                                              \\ \hline
\multicolumn{9}{r||}{\textbf{Total:}}        & \multicolumn{1}{c|}{1316268}           & \multicolumn{1}{c|}{51.23}        & 641884                                                       \\ \cline{9-12}
\end{tabular}
\end{adjustbox}
\caption{IndicHG Analysis: Showing overall duplication and overlap(or data-contamination) percentages.}
\label{table:IndicHG_data_contamination_statistics}
\end{table*}

\subsection{Quantitative Analysis} 
We initiate the analysis by implementing preprocessing steps for the IndicHG dataset, including checks for prefixes, duplicates, and minimum length. In addition to the eight languages we are focusing on, we extended the preprocessing to include the remaining three languages of IndicHG: Oriya, Punjabi, and Assamese.

Surprisingly, despite claims to the contrary, our analysis reveals that the IndicHG dataset contains a significant number of duplicate article-headline pairs in the training, development, and test splits for most languages. Out of the total 1.31 million pairs, approximately 0.67 million (51.23\%) are duplicates.
Moreover, it is ideal for a dataset to have no overlap or common samples among the training, development, and test splits. However, the statistics presented in Table \ref{table:IndicHG_data_contamination_statistics} demonstrate a high level of overlap among these splits for most of the languages, corroborating data contamination. For instance, an article-headline pair\footnote{ \url{https://tinyurl.com/2p85mayt}} from the Kannada language appears 115 times in the training data, 18 times in the development data, and 2 times in the test data.


\begin{figure}[h]
    \centering
    \includegraphics[scale=0.50]{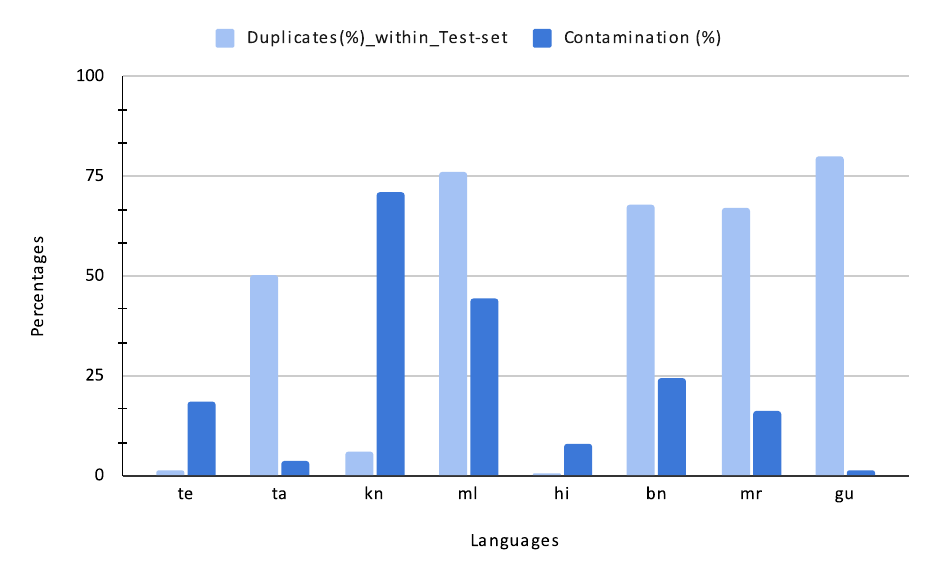}
    \caption{Language-wise data bias in IndicHG test-set.}
    \label{fig:test-bias}
\end{figure}

Data contamination introduces bias in evaluation, as the metrics calculated on the development and test datasets do not accurately represent the model's performance on unseen data. Additionally, we assert that the heavy presence of duplicated data in the dataset may lead models trained on this data to achieve artificially high performance by memorizing the duplicated pairs, thereby hindering their ability to generalize to new, unseen data.

To support our arguments, we take several steps. Firstly, we eliminate all duplicate pairs from each of the training, development, and test splits of the IndicHG dataset. To deal with data contamination, the following 2 variations were attempted: 
\begin{enumerate}[leftmargin=*]
    \item To ensure the integrity of the test set, a straightforward approach was adopted, which involved excluding any pairs that were already present in the corresponding train/dev sets. Additionally, any pairs in the dev set that were already present in the train set were also removed. This approach effectively eliminated data contamination and allowed the training set to remain as large as possible. These splits were then utilized to reproduce the IndicHG results as \textit{IndicHG\_Unbiased}. Notably, this dataset exhibited a significant decrease in average R-L score, with a decrease of 22.48 compared to the score reported in the original IndicNLG paper \citep{13.IndicNLG}, resulting in an average R-L score of 23.42; as outlined in Table \ref{table:Comparison_reproducedIndicNLG}.

    To evaluate the specific impact of data contamination, we divided the IndicHG test set into two subsets. The first subset consisted of pairs from the IndicHG test set that were also present in the corresponding train or dev sets. The second subset comprised the remaining (unique) pairs from the original test set. Figure \ref{fig:RL_linegraph} shows the R-L score comparison\footnote{For details refer to Table \ref{table:overlap-rouge-comparision}} for these two test subsets, referred to as \textit{Overlaps} and \textit{Without\_Overlap} respectively, against those obtained from the total (original) test set. 
    The results unequivocally support the claim that data contamination indeed leads to artificial high performance. 


    \begin{figure}[h]
    \centering
    \includegraphics[scale=0.50]{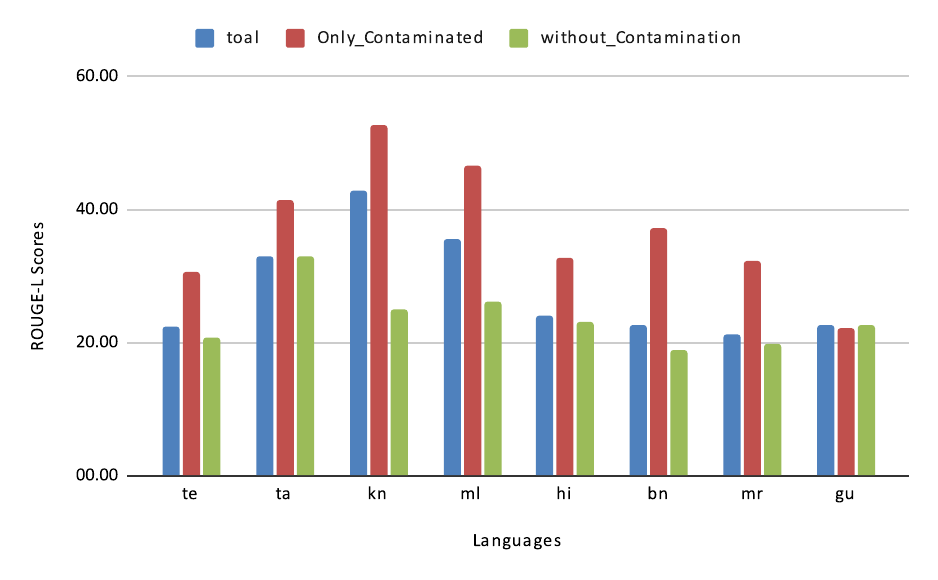}
    \caption{ROUGE-L scores for subsets of IndicHG Test set.}
    \label{fig:RL_linegraph}
    \end{figure}

    \item As an alternative approach, pairs present in the training set that also appeared in the corresponding dev and test sets were eliminated. Similarly, pairs in the dev set that were already present in the test set were excluded. Additionally, pairs were filtered out if the headline was found in the article's prefix, or if the pairs were too short. This method aimed to ensure that the new test set closely resembled the original set while eliminating problematic cases. 
    The stepwise statistics of this filteration process and final split counts are provided in Table \ref{table:IndicHG_filtered_creation_Statistics}. Further statistics of the resulting filtered dataset, referred to as \textit{IndicHG\_filtered}, can be found in Table \ref{table:IndicHG_Filtered_dataset_stats}. 
\end{enumerate} 

While it may seem intuitive that a larger training set would lead to better model training, our findings suggest that both of the aforementioned approaches yield similar scores. Consequently, we have decided to utilize the \textit{IndicHG\_filtered} version for all future cross-comparisons. This is primarily because its test set bears closer resemblance to the original test set. Section \ref{sec:experiment_analysis} describes further experimentation conducted using this dataset.

\subsection{Qualitative Analysis:} 
To conduct a qualitative analysis, we begin by manually evaluating a random selection of article-headline pairs from the IndicHG Telugu dataset\footnote{Manual evaluation was restricted to Telugu, due to limited language experts/resources.}. This dataset comprises articles collected from approximately 22 different Telugu news websites. To ensure a comprehensive evaluation, we assess at least five random pairs from each website. Our evaluation brings to light certain issues that indicate a lack of site-specific scraping implementation in IndicHG. The identified issues are as follows:

\begin{enumerate}
    \label{sec:problems}
    \item Unwanted information (noise) is present at the beginning of the article.

    \item Headline is out of the context of the article.
    
    \item The article part of a pair, itself contains multiple other article-headline pairs.
\end{enumerate}

These quality issues in the article-headline pairs can significantly impact the performance of models. When the headline is contextually unrelated to the article, the generated headlines by the model are inaccurate, resulting in subpar performance. Likewise, the presence of multiple articles within a single article introduces irrelevant information, causing the model to focus on only a fraction of the total content.

For each of the aforementioned issues, we meticulously document the corresponding source website. Subsequently, we employ simple scripts, regular expressions, and other techniques to further examine all the article-headline pairs from these source websites. Among all the issues observed, the most prevalent is the occurrence of multiple articles within a single article (issue-3). By employing basic regular expressions, we were able to detect a total of 5773 such pairs, although not capturing all instances, primarily sourced from the Andhra Bhoomi website\footnote{\url{http://www.andhrabhoomi.net/}}, which constitutes 30\% of the Telugu IndicHG dataset. 
Considering the significant quantity of such pairs, we further update our \textit{IndicHG\_filtered} dataset by eliminating these pairs.

For further examples and additional details regarding all other identified problematic cases, please refer to Appendix \ref{appendix:C2}.

\begin{table*}
\begin{adjustbox}{width=\textwidth, center}
\begin{tabular}{lrrrr|rrrr}
\hline
\multicolumn{1}{c}{}                  & \multicolumn{4}{c}{Dravidian language family} & \multicolumn{4}{c}{Indo-Aryan language family} \\  \cline{2-9}
\multicolumn{1}{c}{}                  & te           & ta          & kn          & ml         & hi           & bn           & mr          & gu          \\ \cline{1-9}
Total \# Pairs               & 26717        & 75954       & 155057      & 20966       & 297284       & 142731       & 142590      & 262457      \\
\# Duplicates                         & 2772         & 38958       & 135693      & 11276       & 13638        & 100835       & 99433       & 200334      \\ 
\textbf{\# Pairs after deduplication} & 23945        & 36996       & 19364       & 9690        & 283646       & 41896        & 43157       & 62123       \\ \hline    
\# Pairs with prefix                  & 669          & 796         & 5           & 22          & 19           & 336          & 6           & 92          \\
\# Pairs with multiple-articles       & 5773         & 0           & 0           & 0           & 0            & 0            & 0           & 0           \\
\# Pairs too short                    & 30           & 5           & 7           & 8           & 1470         & 5            & 8           & 7           \\
\textbf{\# Pairs after filtering}     & 17473        & 36195       & 19352       & 9660        & 282157       & 41555        & 43143       & 62024       \\ \hline    
\# Pairs in train                     & 13539        & 28750       & 13602       & 7235        & 194627       & 32435        & 33772       & 49566       \\
\# Pairs in dev                       & 1903         & 3702        & 2693        & 1177        & 43604        & 4480         & 4644        & 6228        \\
\# Pairs in test                      & 2031         & 3743        & 3057        & 1248        & 43926        & 4640         & 4727        & 6230   
    \\ \hline    
\end{tabular}
\end{adjustbox}
\caption{IndicHG\_filtered dataset creation statistics.}
\label{table:IndicHG_filtered_creation_Statistics}
\end{table*}

\begin{table*}[h]
\centering
\begin{adjustbox}{width=\textwidth, center}
\begin{tabular}{c|c|cccccccc|c}
\hline
\multirow{2}{*}{\textbf{Test sets}} & \multicolumn{1}{c|}{\multirow{2}{*}{\textbf{Models Fine-tuned on}}} & \multicolumn{8}{c|}{\textbf{Language}}                                                                                                                                                                           & \multirow{2}{*}{\textbf{Average}} \\ \cline{3-10}
                                        & \multicolumn{1}{c|}{}                                    &  te     &  ta     &  kn     &  ml     &  hi     &  bn     &  mr     & gu    &                                   \\ \hline 


                                        
\multirow{4}{*}{IndicHG}               
                                       & IndicHG\_filtered                                                  &  17.80   &  33.46  &  22.98  &  25.09  &  24.52  &  19.18  &  21.35  & 22.87 & 23.41                             \\ 
                                        & Mukhyansh                                                  &  \textbf{27.05}  &  \textbf{35.05}  &  \textbf{28.20}   &  \textbf{29.23}  &  \textbf{26.84}  &  17.65  &  \textbf{26.31}  & 19.86 & \textbf{26.27}                             \\ 
                                        & Mukhyansh\_small                                                  &  21.46  &  30.68  &  23.09  &  24.04  &  23.39  &  15.66  &  22.22  & 18.59 & 22.39                             \\ 
                                        \hline

\multirow{5}{*}{IndicHG\_filtered}       
                                        & \cellcolor{gray!25} IndicHG* (with overlap)                             & \cellcolor{gray!25} 20.58  & \cellcolor{gray!25}  32.50   & \cellcolor{gray!25} 41.89  &\cellcolor{gray!25}  33.29  &\cellcolor{gray!25}  23.92  &\cellcolor{gray!25}  21.78  &\cellcolor{gray!25}  21.93  &\cellcolor{gray!25} 22.51 &\cellcolor{gray!25} 27.30 \\ 
                                        
                                        & IndicHG\_filtered                                        &  16.66  &  32.85  &  22.91  &  25.11  &  24.61  &  18.97  &  21.38  & 22.86 & 23.17                             \\ 
                                        & Mukhyansh                                        &  \textbf{24.00}     &  \textbf{34.96}  &  \textbf{27.98}  &  \textbf{29.28}  &  \textbf{26.95}  &  17.66  &  \textbf{26.33} & 19.85 & \textbf{25.88}                              \\ 
                                        & Mukhyansh\_small                                        &  19.67  &  30.54  &  22.98  &  24.00     &  23.50   &  15.66  &  22.24  & 18.59 & 22.15                                 \\ 
                                        \hline    
\multirow{5}{*}{Mukhyansh}       
                                        & IndicHG*                                                &  19.83  &  29.31  &  20.61  &  19.51  &  23.95  &  14.80   &  15.29  & 16.19 & 19.94                             \\  
                                        & IndicHG\_filtered                                                &  17.53  &  29.43  &  18.66  &  20.44  &  26.07  &  16.13  &  15.73  & 16.56 & 20.07                             \\ 
                                        & Mukhyansh                                                &  \textbf{37.33}  &  \textbf{41.16}  &  \textbf{32.59}  &  \textbf{32.04}  &  \textbf{36.18}  &  \textbf{22.04}  &  \textbf{27.08}  & \textbf{23.05} & \textbf{31.43}                             \\
                                        & Mukhyansh\_small                                                &  28.66  &  36.01  &  26.11  &  26.73  &  32.39  &  18.96  &  22.59  & 20.49 & 26.49                                 \\ 
                                        \hline 
\end{tabular} 
\end{adjustbox}
\caption{Performance comparison (by ROUGE-L) of various models.}
\label{tab:model-results-comparision}
\end{table*}

\subsection{Experiments and Analysis}
\label{sec:experiment_analysis}
In order to assess the effectiveness of different models, we fine-tune the SSIB model\footnote{Unless otherwise stated, all experiments conducted in this study were based on the SSIB model.} using a range of specifically crafted training and test sets:
\begin{enumerate}[leftmargin=*]
    \item First, we fine-tune a model on the IndicHG\_filtered dataset and evaluate its performance on the corresponding filtered test set, while ensuring that the fine-tuning hyperparameters remain consistent with those described in the IndicNLG paper.
    The results, as presented in Table \ref{tab:model-results-comparision}, demonstrate the true performance of IndicHG when only good quality unique pairs are considered. It is evident that the ROUGE-L scores decrease significantly compared to the scores produced by the biased data (i.e. unfiltered IndicHG). 
    Next,  other models were also tested on IndicHG\_filtered test set. See, Table \ref{tab:model-results-comparision}. Notably, while testing IndicHG* model on IndicHG\_filtered test set, we are bound to get biased (high) scores. This is because in case of IndicHG\_filtered, the training set itself was prepared without overlapping pairs (leaving them intact in the corresponding test set). Keeping this bias aside, our Mukhyansh model outperforms all the others.

    \item 
    To further investigate the impact of quality vs. quantity, we prepare a smaller version of the Mukhyansh dataset. In order to create the new train, dev, and test sets, separate random sampling is performed over the original train, dev, and test sets of Mukhyansh. A model, called \textit{Mukhyansh\_small}, is then fine-tuned only on this smaller train set, and tested against other models, see Table \ref{tab:model-results-comparision}.
    
\end{enumerate}    

This cross-comparison was then concluded by testing Mukhyansh's SSIB baseline against all other test sets. And as evident by the R-L scores (highlighted as bold) in Table \ref{tab:model-results-comparision} Mukhyansh outperforms almost all the other models.

We acknowledge the multilingual models as the limitation and future scope of this work. Due to limited compute-resources we could not fine-tune any multilingual models. However, we believe that multilingual fine-tuning on Mukhyansh dataset would give new state-of-the-art models.

\section{Conclusion}
\label{sec:conclusion}
Headline generation in low-resource languages, such as Indian languages, faces significant challenges due to the scarcity of large, high-quality annotated data. Our work address this gap by introducing Mukhyansh, a comprehensive multilingual dataset comprising over 3.39 million article-headline pairs across eight prominent Indian languages. The importance of our work is substantiated by empirical analysis of existing works, uncovering critical data quality issues. Through extensive experimentation, we demonstrate the superiority of Mukhyansh and our SSIB baseline model, surpassing all existing works in Indian language headline generation.
This achievement highlights the effectiveness of Mukhyansh in advancing research efforts in low-resource language processing and establishes it as a valuable resource for future exploration and innovation in this field.

\section{Ethics Statement} The distribution of the dataset collected from the web raises ethical considerations. We acknowledge that the copyright of the news articles collected from various websites remains with the original creators. Considering that each website may have its own policies regarding data distribution or public availability, we offer researchers the URLs and web scraping scripts necessary to reproduce the data, ensuring transparency and encouraging proper attribution through the release of the list of URLs under the Creative Commons license\footnote{\url{https://creativecommons.org/licenses/by/4.0/}}. To ensure the reproducibility of the model results, we plan to release various baseline model checkpoints used for headline generation at a later date.

\bibliography{acl2023}
\bibliographystyle{acl_natbib}

\appendix
\newpage
\label{appendix}

\section{Mukhyansh Dataset Additional Details}
\label{appendix:A}
\begin{itemize}
\item Websites used for scraping: To make the dataset more diverse, the data is scraped from a total of 47 websites across all 8 languages, and the list of websites is provided in Table \ref{table: List of websites used for web scraping to collect the data}.
\item To eliminate any bias towards particular news categories we make sure that the scraped dataset covers diverse set of news categories. The category/domain-wise statistics of the Mukhyansh dataset are presented in Table \ref{table:Category wise statistics of Mukhyansh}.
\item To evaluate the task's abstractive nature and difficulty, we compute the percentage of novel n-grams and employ extractive baselines like LEAD-1 and EXT-ORACLE ROUGE-L (R-L) scores. The "percentage of novel n-grams" indicates the proportion of n-grams present in the headline but not found in the article, quantifying the level of uniqueness in the generated summary. Specifically, LEAD-1 R-L calculates the similarity between the first sentence of the article and the reference headline, while EXT-ORACLE R-L computes scores by selecting the sentence from the article that achieves the highest R-L scores with the reference headline. The resulting scores along with other statistics are detailed in Table \ref{tab:mukhyansh_detailed_stats}

\end{itemize}

\begin{table*}[h]
\begin{adjustbox}{width=\textwidth, center}
\begin{tabular}{|c|l|l|c|l|l|c|l|l|}
\hline
S.No & {L}  & \multicolumn{1}{c|}{Website}      & S.No & L  & \multicolumn{1}{c|}{Website}        & S.No & L  & \multicolumn{1}{c|}{Website}        \\
\hline
1    & te & https://www.ap7am.com/telugu-news & 17   & kn & https://kannadanewsnow.com/kannada/ & 33   & ml & https://eveningkerala.com/          \\
2    & te & https://www.prabhanews.com/       & 18   & kn & https://hosadigantha.com/           & 34   & hi & https://www.jagran.com/             \\
3    & te & https://www.suryaa.com/index.html & 19   & kn & https://kannada.asianetnews.com/    & 35   & hi & https://www.khaskhabar.com/         \\
4    & te & https://www.manatelangana.news/   & 20   & kn & https://newskannada.com/            & 36   & hi & https://www.indiatv.in/             \\
5    & te & http://www.andhrabhoomi.net/      & 21   & kn & https://www.kannadaprabha.com/      & 37   & bn & https://www.anandabazar.com/        \\
6    & te & https://prajasakti.com/           & 22   & kn & https://www.sahilonline.net/ka      & 38   & bn & https://www.sangbadpratidin.in/     \\
7    & te & https://www.vaartha.com/          & 23   & kn & https://www.udayavani.com/          & 39   & bn & https://bengali.abplive.com/live-tv \\
8    & te & https://10tv.in/                  & 24   & kn & http://vishwavani.news/             & 40   & bn & https://uttarbangasambad.com/       \\
9    & te & https://www.hmtvlive.com/         & 25   & kn & https://ainlivenews.com/            & 41   & bn & https://bangla.asianetnews.com/     \\
10   & ta & https://www.hindutamil.in/        & 26   & kn & https://vaarte.com/                 & 42   & mr & https://www.lokmat.com/             \\
11   & ta & https://www.polimernews.com/      & 27   & kn & https://btvkannada.com/             & 43   & mr & https://prahaar.in/                 \\
12   & ta & https://tamil.asianetnews.com/    & 28   & ml & https://www.eastcoastdaily.com/     & 44   & mr & https://marathi.abplive.com/        \\
13   & ta & https://www.updatenews360.com/    & 29   & ml & https://suprabhaatham.com/          & 45   & gu & https://sandesh.com/                \\
14   & kn & https://kannadadunia.com/         & 30   & ml & https://www.bignewslive.com/        & 46   & gu & https://www.gujaratsamachar.com/    \\
15   & kn & https://eesanje.com/              & 31   & ml & https://www.malayalamexpress.in/    & 47   & gu & https://gujarati.news18.com/        \\
16   & kn & https://www.vijayavani.net/       & 32   & ml & https://dailyindianherald.com/      &      &    &                                     \\                
\hline
\end{tabular}
\end{adjustbox}
\caption{List of websites used for creating Mukhyansh.}
\label{table: List of websites used for web scraping to collect the data}
\end{table*}

\newpage

\begin{table*}
\centering
\begin{tabular}{c|rrrrrrrr} \hline
\multirow{2}{*}{\begin{tabular}[c]{@{}c@{}}News Category\end{tabular}} & \multicolumn{8}{c}{Category-wise counts of article-headline pairs for each language} \\ \cline{2-9}
                                                                                  & te        & ta        & kn        & ml        & hi        & bn        & mr        & gu        \\ \hline
state                                                                             & 698059    & 133599    & 163857    & 144491    &  -       & 143804    & 184045    & 123183    \\
national                                                                          & 91787     & 80711     & 61170     & 92833     & 314528    & 42913     & 72182     & 53248     \\
entertainment                                                                     & 59244     & 31265     & 22697     & 14939     & 80202     & 31470     & 2819      & 19710     \\
international                                                                     & 24262     & 29463     & 26092     & 34008     & 29668     & 20552     & 15347     & 37682     \\
sports                                                                            & 19933     & 26186     & 18775     & 10204     & 78190     & 30676     & 29947     & 19337     \\
business                                                                          & 13495     & 12874     & 8747      & 3446      & 60524     & 775       & 10379     & 21884     \\
crime                                                                             & 8917      & 6656      & 7541      & 7064      & 8052      &  -         & 16489     &   -        \\
covid                                                                             & 1425      & 6470      & 14147     & 4348      &    -       & 4205      &     -      &   -        \\
politics                                                                          &  -         & 4484      & 5816      & 843       & 29459     & 346       & 3234      &     -      \\
other                                                                             &  -         &   -        & 9081      & 2896      &   -        & 6532      &   -        & 914 \\ \hline
\end{tabular}

\caption{Category wise statistics of Mukhyansh}
\label{table:Category wise statistics of Mukhyansh}
\end{table*}

\begin{table*}[]
\resizebox{\textwidth}{!}{
\begin{tabular}{clrccrcrccccccc}
\hline
\multirow{2}{*}{L} & \multicolumn{1}{c}{\multirow{2}{*}{\begin{tabular}[c]{@{}c@{}}Total \\ Pairs\end{tabular}}} & \multirow{2}{*}{\begin{tabular}[c]{@{}c@{}}Avg sents\\ in article\end{tabular}} & \multirow{2}{*}{\begin{tabular}[c]{@{}c@{}}Avg tokens\\ in article\end{tabular}} & \multirow{2}{*}{\begin{tabular}[c]{@{}c@{}}Avg tokens\\ in headline\end{tabular}} & \multicolumn{2}{c}{Total Tokens} & \multicolumn{2}{c}{Unique Tokens} & \multicolumn{4}{c}{\% novel n-gram} & Lead-1 & \begin{tabular}[c]{@{}c@{}}EXT-\\ORACLE\end{tabular} \\
                            & \multicolumn{1}{c}{}                                                                         &                                                                                 &                                                                                  &                                                                                   & articles       & headlines       & articles        & headlines       & n=1     & n=2    & n=3    & n=4    & R-L    & R-L                                                   \\ \hline
te                          & 917122                                                                                       & \phantom{0}7.97                                                                            & 103.64                                                                           & \phantom{0}7.42                                                                              & 95.05M        & 6.80M          & 2.3M         & 376K          & 36.63   & 62.87  & 82.10   & 91.41  & 23.54  & 33.21                                                 \\
ta                          & 331708                                                                                       & 15.47                                                                           & 218.99                                                                           & 11.50                                                                              & 72.64M        & 3.82M          & 1.8M         & 225K          & 33.02   & 55.12  & 73.75  & 85.05  & 32.70   & 39.33                                                 \\
kn                          & 337923                                                                                       & 10.94                                                                           & 154.77                                                                           & \phantom{0}9.03                                                                              & 52.3M         & 3.05M          & 1.9M         & 222K          & 41.30    & 65.88  & 82.73  & 91.45  & 19.66  & 30.08                                                 \\
ml                          & 315072                                                                                       & 10.26                                                                           & 115.45                                                                           & \phantom{0}9.54                                                                              & 36.37M        & 3.01M          & 2.5M         & 351K          & 36.14   & 55.59  & 71.20   & 81.73  & 34.60   & 41.94                                                 \\
hi                          & 600623                                                                                       & 14.54                                                                           & 303.05                                                                           & 13.45                                                                             & 182.02M       & 8.08M          & 1.3M         & 137K          & 20.31   & 47.20   & 67.96  & 81.27  & 25.99  & 35.02                                                 \\
bn                          & 281273                                                                                       & 19.41                                                                           & 244.78                                                                           & 10.10                                                                              & 68.85M        & 2.84M          & 0.9M          & 135K          & 37.60    & 67.60   & 84.31  & 92.27  & 15.51  & 30.50                                                  \\
mr                          & 334442                                                                                       & 17.71                                                                           & 271.02                                                                           & \phantom{0}8.41                                                                              & 90.64M        & 2.81M          & 1.9M         & 241K          & 37.11   & 64.73  & 82.66  & 91.66  & 13.88  & 28.34                                                 \\
gu                          & 275958                                                                                       & 16.45                                                                           & 284.39                                                                           & 12.46                                                                             & 78.48M        & 3.44M          & 1.7M         & 197K          & 38.24   & 65.81  & 82.08  & 90.54  & 12.21  & 28.72
\\ \hline
\end{tabular}
}
\caption{Mukhyansh dataset statistics in detail.}
\label{tab:mukhyansh_detailed_stats}
\end{table*}

\begin{table*}[]
\resizebox{\textwidth}{!}{
\begin{tabular}{crcccrrcrcccccc}
\hline
\multirow{2}{*}{L} & \multirow{2}{*}{\begin{tabular}[c]{@{}c@{}}Total\\ Pairs\end{tabular}} & \multirow{2}{*}{\begin{tabular}[c]{@{}c@{}}Avg sents\\ in article\end{tabular}} & \multirow{2}{*}{\begin{tabular}[c]{@{}c@{}}Avg tokens \\ in article\end{tabular}} & \multirow{2}{*}{\begin{tabular}[c]{@{}c@{}}Avg tokens \\ in headline\end{tabular}} & \multicolumn{2}{c}{Total Tokens} & \multicolumn{2}{c}{Unique Tokens} & \multicolumn{4}{c}{\% novel n-gram} & Lead-1 & \begin{tabular}[c]{@{}c@{}}EXT-\\ ORACLE\end{tabular} \\
                   &                                                                        &                                                                                 &                                                                                   &                                                                                    & articles      & titles     & articles         & titles         & n=1     & n=2     & n=3    & n=4    & R-L    & R-L                                                   \\ \hline
te                 & 17473                                                                  & 13.99                                                                           & 185.09                                                                            & \phantom{0}7.97                                                                               & 3.2M          & 139K        & 238K           & 35.6K         & 36.26   & 65.66   & 85.87  & 94.08  & 15.23  & 29.30                                                  \\
ta                 & 36195                                                                  & 13.43                                                                           & 181.77                                                                            & 11.76                                                                              & 6.58M          & 425K        & 311K           & 51.9K          & 32.94   & 54.89   & 70.17  & 78.96  & 33.46  & 40.10                                                  \\
kn                 & 19352                                                                  & 11.49                                                                           & 189.16                                                                            & \phantom{0}9.22                                                                               & 3.66M          & 178K        & 237K           & 34.1K          & 33.02   & 57.47   & 75.89  & 86.59  & 18.19  & 29.73                                                 \\
ml                 & 9660                                                                   & 13.55                                                                           & 168.38                                                                            & 10.08                                                                              & 1.63M          & \phantom{0}97K         & 232K           & 29K          & 39.15   & 61.48   & 77.78  & 87.04  & 26.40   & 35.79                                                 \\
hi                 & 282157                                                                 & 18.25                                                                           & 397.08                                                                            & 12.55                                                                              & 112.04M        & 3.5M       & 543K           & 74.9K          & 20.79   & 49.86   & 71.06  & 83.56  & 21.99  & 32.52                                                 \\
bn                 & 41555                                                                  & 14.65                                                                           & 239.55                                                                            & 11.27                                                                              & 9.95M          & 468K        & 245K           & 47.2K          & 38.02   & 64.35   & 80.91  & 89.33  & 13.95  & 27.39                                                 \\
mr                 & 43143                                                                  & 13.61                                                                           & 205.31                                                                            & \phantom{0}8.57                                                                               & 8.86M          & 369K        & 258K           & 51K          & 31.45   & 57.76   & 77.44  & 86.59  & 13.08  & 32.55                                                 \\
gu                 & 62024                                                                  & 12.31                                                                           & 226.64                                                                            & 11.20                                                                               & 14.06M         & 694K        & 425K           & 81.5K          & 35.85   & 60.69   & 76.75  & 85.94  & 15.52  & 29.39     
\\ \hline
\end{tabular}
}
\caption{IndicHG\_filtered dataset statistics in detail.}
\label{table:IndicHG_Filtered_dataset_stats}
\end{table*}

\section{IndicHG Analysis}
\label{appendix:B}
\subsection{Reproduced Results }
In this section, we present the results of the experiment conducted to reproduce the results in the IndicNLG paper by fine-tuning the SSIB model on the IndicHG dataset. We report the mean and standard deviation of R-1, R-2, and R-L scores across multiple runs (i.e. using 5 different seeds to initialize the model). Table \ref{table:Means_Reproduced_IndicHG_Dataset} provides the detailed statistics.

\begin{table*}
\centering
\begin{tabular}{c|cc|cc|cc}
\hline
\multirow{2}{*}{L} & \multicolumn{2}{c|}{R-1}                 & \multicolumn{2}{c|}{R-2}                 & \multicolumn{2}{c}{R-L}                 \\ \cline{2-7} 
                            & \multicolumn{1}{c|}{mean} & std & \multicolumn{1}{c|}{mean} & std & \multicolumn{1}{c|}{mean} & std \\ \hline
te                          & \multicolumn{1}{c|}{23.75}         & 1.31         & \multicolumn{1}{c|}{11.98}         & 0.88         & \multicolumn{1}{c|}{22.37}         & 1.28         \\ 
ta                          & \multicolumn{1}{c|}{34.49}         & 0.70          & \multicolumn{1}{c|}{21.06}         & 0.62         & \multicolumn{1}{c|}{32.96}         & 0.74         \\ 
kn                          & \multicolumn{1}{c|}{43.85}         & 1.41         & \multicolumn{1}{c|}{35.89}         & 1.58         & \multicolumn{1}{c|}{42.79}         & 1.43         \\ 
ml                          & \multicolumn{1}{c|}{37.20}          & 1.62         & \multicolumn{1}{c|}{25.59}         & 1.89         & \multicolumn{1}{c|}{35.64}         & 1.72         \\ 
hi                          & \multicolumn{1}{c|}{28.73}         & 0.63         & \multicolumn{1}{c|}{13.42}         & 0.37         & \multicolumn{1}{c|}{24.12}         & 0.62         \\ 
bn                          & \multicolumn{1}{c|}{24.54}         & 0.29         & \multicolumn{1}{c|}{12.58}         & 0.36         & \multicolumn{1}{c|}{22.54}         & 0.31         \\ 
mr                          & \multicolumn{1}{c|}{22.99}         & 0.46         & \multicolumn{1}{c|}{11.26}         & 0.28         & \multicolumn{1}{c|}{21.28}         & 0.38         \\ 
gu                          & \multicolumn{1}{c|}{24.77}         & 0.22         & \multicolumn{1}{c|}{11.87}         & 0.15         & \multicolumn{1}{c|}{22.68}         & 0.35         \\ \hline
Average            & \multicolumn{1}{c|}{30.04}         & 0.83         & \multicolumn{1}{c|}{17.96}         & 0.77         & \multicolumn{1}{c|}{28.05}         & 0.85         \\ \hline
\end{tabular}
\caption{Mean \& Standard Deviation of 5 iterations of IndicHG* results}
\label{table:Means_Reproduced_IndicHG_Dataset}
\end{table*}

\begin{table*}[h!]
\centering
\begin{adjustbox}{width=\textwidth, center}
\begin{tabular}{c|c|cccccccc|c}
\hline
\multirow{2}{*}{\textbf{Test sets}} & \multicolumn{1}{c|}{\multirow{2}{*}{\textbf{Models Fine-tuned on}}} & \multicolumn{8}{c|}{\textbf{Language}}                                                                                                                                                                           & \multirow{2}{*}{\textbf{Average}} \\ \cline{3-10}
                                        & \multicolumn{1}{c|}{}                                    &  te     &  ta     &  kn     &  ml     &  hi     &  bn     &  mr     & gu    &                                   \\ \hline

\multirow{1}{*}{IndicHG}               
                                        & \multirow{3}{*}{IndicHG*}                                  & 22.37 & 32.96 & 42.79 & 35.64 & 24.12 & 22.54 & 21.28 & 22.68 & 28.05                             \\ 
\multirow{1}{*} {Overlaps (IndicHG)} 
                                        &                 &  30.53  &  41.36   &  52.63  &  46.61  &  32.81  &  37.12  &  32.21  & 22.27 & 36.94                      \\ 

\multirow{1}{*}{IndicHG-Overlaps}
                                        &                 &  20.84  &  32.87   &  25.00  &  26.07  &  23.08  &  18.79  & 19.92  & 22.53 & 23.64                      \\     \hline
                                        
\end{tabular} 
\end{adjustbox}
\caption{Impact of Overlap on  IndicHG Performance (by ROUGE-L).}
\label{table:overlap-rouge-comparision}
\end{table*}

\subsection{Problem Cases}
\label{appendix:C2}
\begin{figure}[h]
    \centering
    \includegraphics[scale=0.50]{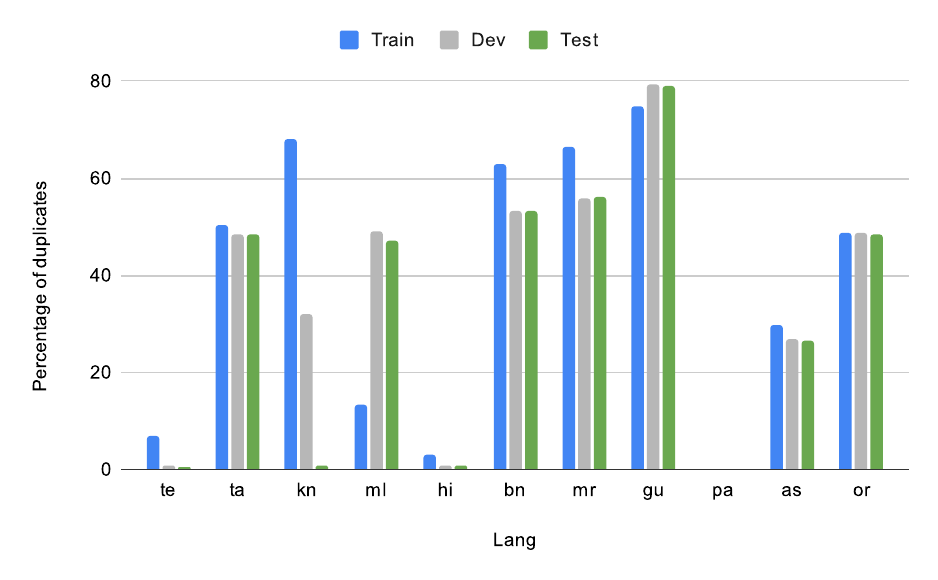}
    \caption{Duplication percentage within IndicHG train,dev,test splits.}
    \label{fig:split-wise-duplicates}
\end{figure}

In this section we present various issues that are present in IndicHG dataset. Table \ref{table:overlap-rouge-comparision} gives language-wise ROUGE-L scores for overlapping and non-overlapping pairs of the IndicHG test set against the scores of the total test set. Figure \ref{fig:split-wise-duplicates} dipicts the percentages of duplication remained in train, dev and test splits of IndicHG after removing all overlapping pairs.


An example of the prefix case is presented in Table \ref{table:example_prefix}. 

Sample article-headline pairs pertaining to the issues mentioned in section \ref{sec:problems} are presented in Table \ref{table:example_outofcontext} and Table \ref{table:example_multiple articles}. For better readability, instead of the Telugu script, we transliterate the text into Latin characters using ISO 15919 standard code.

The analysis of article-headline pairs of BBC Telugu\footnote{\url{https://www.bbc.com/telugu}}, BBC Tamil\footnote{\url{https://www.bbc.com/tamil}} websites that are present in IndicHG dataset is detailed in Table \ref{table: BBC Errors}.


\begin{table*}[h!]
\begin{adjustbox}{width=\textwidth, center}
\begin{tabular}{|l|}
\hline
\textbf{URL:} \url{https://www.bbc.com/telugu/india-48363611}  \\ \hline                                                                                                                                            
\textbf{Headline:} vaisīpī mèjāriṭīki prajāśāṃti pārṭī gaṃḍikòṭṭiṃdā? òke peruto nilabèṭṭina abhyarthulaku vaccina oṭlènni? - BBC News tèlugu    \\ \hline                                                                                                                                                    
\textbf{Article:} \colorbox{cyan}{vaisīpī mèjāriṭīki prajāśāṃti pārṭī gaṃḍikòṭṭiṃdā? òke peruto nilabèṭṭina abhyarthulaku vaccina oṭlènni? }\colorbox{pink}{24 me 2019 dīnini }\\\colorbox{pink}{kriṃdi vāṭito ṣer ceyaṃḍi ivi bayaṭi liṃklu, kābaṭṭi kòtta viṃḍolo tèravabaḍatāyi ivi bayaṭi liṃklu, kābaṭṭi kòtta viṃḍolo tèravabaḍatāyi ṣer}\\ \colorbox{pink}{pyānèlnu mūsiveyaṃḍi} āṃdhapradeś ènnikallo kee pāl netṛtvaṃloni prajāśāṃti pārṭī cālā coṭla tana abhyarthulanu bariloki diṃpiṃdi.\\ kònni coṭla vaisīpī abhyarthula perlanu polina vyakthulanu bariloki diṃpiṃdane vārtalu vaccāyi. dīnipai vaisīpī pratinidhulu mārci 26na \\ dillīki vacci ènnikala saṃghāniki phiryādu kūḍā ceśāru.dādāpu 35 niyojakavargāllo tama abhyarthulanu polina abhyarthulanu prajāśāṃti \\ poṭīlo nilabèṭṭiṃdani, dīnipai caryalu tīsukovālani koriṃdi.prajāśāṃti ènnikala gurtu ayina hèlikāpṭar kūḍā tama phyān gurtunu poli uṃdani, \\ dīnipainā caryalu tīsukovālani koriṃdi.ayite, kee pāl nilabèṭṭina abhyarthula valla vaisīpīki naṣṭaṃ jarigiṃdā..? e niyojakavargāllo vaisīpī \\ abhyarthula mèjāriṭīpai prabhāvaṃ paḍiṃdi? phalitālu èlā unnāyi? anedi kiṃdi paṭṭikalo cūḍòccu.kramasaṃkhya \\ \hline
                                                                                                                                   
\end{tabular}
\end{adjustbox}
\caption{Example  of a headline that is directly present in article's prefix. The text highlighted in cyan color is the prefix information which is the same as the headline, and the one in pink is  unwanted information (noise).}
\label{table:example_prefix}
\end{table*}


\begin{table*}[h!]
\begin{adjustbox}{width=\textwidth, center}
\begin{tabular}{|l|}
\hline
\textbf{URL:} \url{https://www.bbc.com/telugu/india-42493669} \\ \hline
\textbf{Headline:} daṃgal bāhubali.. rèṃḍū rèṃḍe - BBC News tèlugu \\ \hline                                                                                                                                      
\textbf{Article:} \colorbox{cyan}{insṭèṃṭ ṭripul talākku cèllu }\colorbox{lime}{daśābdālugā èṃto maṃdi musliṃ mahil̤ala vedanaku kāraṇamaina vidhānaṃ..'insṭèṃṭ ṭripul talāk'.ī islāmik ācārānni rājyāṃga } \\ \colorbox{lime}{viruddhamani tīrmānistū deśa atyunnata nyāyasthānaṃ āgasṭulo cāritraka tīrpuni vèluvariṃciṃdi.aiduguru sabhyulunna dharmāsanaṃlo mugguru jaḍjilu 'insṭèṃṭ } \\ \colorbox{lime}{ṭripul talāk' rājyāṃga viruddhamanī , adi mahil̤alapai vivakṣa cūpedigā uṃdanī perkònnāru.suprīṃ korṭu prakaṭiṃcina ī nirṇayaṃ paṭla deśa prajalu, mukhyaṃgā } \\ 
\colorbox{lime}{musliṃ mahil̤alu harṣaṃ vyaktaṃ ceśāru.pārlamèṃṭulo ṭripul talāk billuni saitaṃ praveśa pèṭṭaḍaṃto dāniki saṃbaṃdhiṃcina caṭṭa rūpakalpanalo maro muṃdaḍugu } \\ \colorbox{lime}{paḍiṃdi.kānī kònni musliṃ mahil̤ā saṃghālu, āl iṃḍiyā musliṃ parsanal lā borḍto sahā kònni rājakīya pārṭīlu mātraṃ ā billuni vyatirekistunnāyi.} \colorbox{yellow}{bhārat naṃ.1 }\\ 
\colorbox{lightgray}{èkkuva maṃdi bhāratīyulu iṣṭapaḍe krīḍa krikèṭ.kohlī siks kòṭṭinā, bumrā vikèṭ tīsinā adi tama ghanatenannaṭṭu krīḍābhimānulu saṃbara paḍatāru. alāṃṭi abhimānulanu } \\
\colorbox{lightgray}{utsāha parice maro arudaina mailurāyini bhārata krikèṭ jaṭṭu ī eḍādi tòlisāri namodu cesiṃdi. sèpṭèṃbarulo prakaṭiṃcina aisīsī ryāṃkullo aṭu ṭèsṭulū, iṭu vanḍellonū } \\
\colorbox{lightgray}{bhārat naṃbar.1 sthānānni kaivasaṃ cesukuṃdi. òkesāri ilā rèṃḍu phārmāṭlalo tòli sthānaṃlo nilavaḍaṃ bhārata jaṭṭuki ide mòdaṭisāri.kāgā, rohit śarma ī eḍādi civaralo } \\ \colorbox{lightgray}{vanḍello mūḍo dviśatakaṃ sādhiṃci prapaṃca rikārḍu nèlakòlpāḍu.mithālī sārathyaṃlo bhārata mahil̤ā krikèṭ jaṭṭu prapaṃca kap phainalku ceri rannarapgā niliciṃdi.} \\ \colorbox{lightgray}{phibravarilo jarigina '2017 blaiṃḍ varalḍ ṭī20' krikèṭ ṭornīni kūḍā bhārata aṃdhula krikèṭ jaṭṭe gèlucukuṃdi. maropakka byāḍmiṃṭanlo tèlugu kurrāḍu kidāṃbi śrīkāṃt} \\ \colorbox{lightgray}{kòtta caritra sṛṣṭiṃcāḍu.òka eḍādilo nālugu sūpar sirīs ṭaiṭil̤lu gèlucukunna tòli bhāratīyuḍigā rikārḍu nèlakòlpāḍu. iṃḍònesiyā, āsṭreliyā, ḍènmārk, phrāns deśāllo } \\
\colorbox{lightgray}{jarigina sūpar sirīs ṭornīllo śrīkāṃt vijetagā nilicāḍu} \\ \hline                                                                                                                                            
\end{tabular}
\end{adjustbox}

\caption{Example of a headline that is out of context to the article. The text highlighted in cyan is the headline of the article (highlighted in lime), and the text highlighted in yellow is the headline of the article (highlighted in gray). Here, the actual headline has no context in the article.}
\label{table:example_outofcontext}
\end{table*}

\begin{table*}[h!]
\begin{adjustbox}{width=\textwidth, center}
\begin{tabular}{|l|}
\hline
\textbf{URL:} \url{http://www.andhrabhoomi.net/content/dudddd}  \\ \hline                                                                                                                                                                                                                                                                                                                                                     
\textbf{Headline:} vimānaṃ ṭāyilèṭlo 3 kilola baṃgāraṃ svādhīnaṃ       \\ \hline                                                                                                                                          
\textbf{Article:} muṃbayi: dubāyi nuṃci ikkaḍiki vaccina vimānaṃlo polīsulu sodālu ceyagā ṭāyilèṭlo 3 kilola baṃgāraṃ bayaṭa paḍiṃdi. dubāyi nuṃci vaccina \\ prayāṇikullo èvaro ī baṃgārānni tècci ṭāyilèṭlo vadilesi uṃṭārani polīsulu cèbutunnāru.kasṭams tanikhīllo dòrikipote kesulu pèḍatāranna bhayaṃto ilā \\ baṃgārānni vadilesi uṃṭārani polīsulu anumānistunnāru. \colorbox{cyan}{bhārat èdugudalalo yūpī kīlakaṃ} 
\colorbox{yellow}{lakno: bhārat ayidu ṭriliyan ḍālarla ārthika vyavasthagā} \\ \colorbox{yellow}{avatariṃcaḍaṃlo, 2030 nāṭiki prapaṃcaṃloni mūḍu atyaṃta pèdda ārthika vyavasthalalo òkaṭigā èdagaḍaṃlo uttarpradeś òka mukhyamayina pātra }\\ 
\colorbox{yellow}{poṣistuṃdani rakṣaṇa śākha maṃtri rājnāth siṃg annāru.} \colorbox{cyan}{padi maṃdiki kebinèṭ padavulu}		
\colorbox{yellow}{bèṃgal̤ūru, phibravari 6: karnāṭakalo kāṃgrès- jeḍīès} \\
\colorbox{yellow}{saṃkīrṇa prabhutvānni kuppakūlci bījepī adhikāraṃloki rāvaḍāniki sahakariṃcina 10 maṃdi phirāyiṃpu dārulaku mukhyamaṃtri yèḍyūrappa maṃtri} 	 \\
\colorbox{yellow}{vargaṃlo kebinèṭ padavulu labhiṃcāyi.} \colorbox{cyan}{'iṃṭarnèṭ' prāthamika hakkukādu}
\colorbox{yellow}{nyūḍhillī, phibravari 6: iṃṭarnèṭ viniyogiṃcukune hakku prāthamika hakku} \\ 
\colorbox{yellow}{kādani, adi èṃta mātraṃ deśa bhadratato samānamaina prādhānyatanu kaligi unnadi kādani keṃdra maṃtri raviśaṃkar prasād guruvāraṃ rājyasabhalo} 		\\
\colorbox{yellow}{prakaṭana ceśāru.deśa bhadratā paristhitulanu kūḍā aṃte prādhānyatato pariśīliṃcālsina avasaraṃ uṃdannāru.} \\ \hline
   
\end{tabular}
\end{adjustbox}
\caption{Example of article-headline pair with multiple unrelated articles and headlines present in the same piece of text. The text highlighted in cyan color is the headline, followed by its article highlighted in yellow.}
\label{table:example_multiple articles}
\end{table*}

\section{Examples of Model generated Headlines}
\label{appendix:D}
This section presents the examples of headlines generated by various baseline models fine-tuned on Mukhyansh. Table \ref{table:Hindi Model Genrated Outputs} and Table \ref{table:Telugu Model Generated Outputs} presents Hindi, Telugu examples respectively. 

\begin{table*}[]
\centering
\begin{tabular}{ccc}
\hline
\textbf{Error Cases}         & \textbf{Telugu} & \textbf{Tamil} \\ \hline
\# Pairs                   & 1587                & 3800               \\ 
\# Pairs with headline present in prefix & 484        & 1558         \\ 
\# Pairs with unwanted information in the article & 1390       & 3494      \\ 
\# Pairs with above two issues in common & 461        & 1436      \\ 
\# Pairs with headline that is out of the context to the article & 174        & 184        \\ \hline
\end{tabular}
\caption{Statistics of problematic pairs of IndicHG dataset.}
\label{table: BBC Errors}
\end{table*}

\begin{table*}
\begin{adjustbox}{width=\textwidth, center}
\begin{tabular}{|l|l|}
\hline
\textbf{URL}             & \url{https://www.jagran.com//news/national-five-children-killed-in-wall-collapse-incidents-10655913.html}                                                                                                                                                                                                                                                                                                                                                                                                                                                                                                                                                                                                                                                                                                                                                                                                                                                                                                                                                                                                                                                       \\ \hline
\textbf{Article}         & \begin{tabular}[c]{@{}l@{}}\textbf{Transliteration:}  \\ kauśāṃbī| uttara pradeśa ke kauśāṃbī jile meṃ do alaga-alaga jagahoṃ para divāra girane se pāṃca \\ baccoṃ kī mauta ho gaī | pulisa ne somavāra ko batāyā ki kauśāṃbī jile ke patharāvana gāṃva meṃ \\ ravivāra śāma ko miṭṭī se bane ghara kā divāra acānaka gira gayā। divāra ke girane se subhāṣa, usakī \\ bahana lakṣmī aura kuṃdrā kī dabane se mauta ho gaī | pulisa ne batāyā ki dūsarī ghaṭanā ayānā eriyā \\ ke kārakāpura gāṃva meṃ divāra girane se do bacce rajanīśa aura priyaṃkā kī bhī mauta ghaṭanāsthala \\ para hī ho gaī |\\  \textbf{Translation:} \\ Kaushambi. In Uttar Pradesh's Kaushambi district, five children died due to wall collapse at two different\\  places. Police said on Monday that the wall of a house made of mud suddenly collapsed in Pathravan \\ village of Kaushambi district on Sunday evening. Subhash, his sister Lakshmi and Kundra died due to the \\ collapse of the wall. Police said that in the second incident, two children Rajneesh and Priyanka also died \\ on the spot due to wall collapse in Karkapur village of Ayana area.\end{tabular} \\ \hline
\textbf{Actual Headline} & \begin{tabular}[c]{@{}l@{}}\textbf{Transliteration:} \\ yūpī meṃ divāra girane se pāṃca baccoṃ kī mauta\\ \textbf{Translation:} \\ Five children died due to wall collapse in UP\end{tabular}                                                                                                                                                                                                                                                                                                                                                                                                                                                                                                                                                                                                                                                                                                                                                                                                                                                                                                                                                                               \\ \hline
\textbf{GRU + FastText}  & \begin{tabular}[c]{@{}l@{}}\textbf{Transliteration:} \\ yūpī meṃ do alaga jagahoṃ para dīvāra girane se 5 baccoṃ kī mauta\\ \textbf{Translation:} \\ 5 children died due to wall collapse at two different places in UP\end{tabular}                                                                                                                                                                                                                                                                                                                                                                                                                                                                                                                                                                                                                                                                                                                                                                                                                                                                                                                                        \\ \hline
\textbf{LSTM + FastText} & \begin{tabular}[c]{@{}l@{}}\textbf{Transliteration:} \\ yūpī meṃ do alaga hādasoṃ se pāṃca kī mauta\\ \textbf{Translation:} \\ Five killed in two separate accidents in UP\end{tabular}                                                                                                                                                                                                                                                                                                                                                                                                                                                                                                                                                                                                                                                                                                                                                                                                                                                                                                                                                                                     \\ \hline
\textbf{GRU + BPEmb}     & \begin{tabular}[c]{@{}l@{}}\textbf{Transliteration:} \\ sar̤aka hādase meṃ 0 baccoṃ kī mauta\\ \textbf{Translation:} \\ 0 children died in road accident\end{tabular}                                                                                                                                                                                                                                                                                                                                                                                                                                                                                                                                                                                                                                                                                                                                                                                                                                                                                                                                                                                                       \\ \hline
\textbf{mT5-small}       & \begin{tabular}[c]{@{}l@{}}\textbf{Transliteration:} \\ uttara pradeśa meṃ do jagahoṃ para divāra girane se 5 baccoṃ kī mauta\\ \textbf{Translation:} \\ 5 children died due to wall collapse at two places in Uttar Pradesh\end{tabular}                                                                                                                                                                                                                                                                                                                                                                                                                                                                                                                                                                                                                                                                                                                                                                                                                                                                                                                                   \\ \hline
\textbf{SSIB}            & \begin{tabular}[c]{@{}l@{}}\textbf{Transliteration:} \\ yūpī ke kauśāṃbī meṃ do alaga alaga jagahoṃ para divāra girane se 5 baccoṃ kī mauta\\ \textbf{Translation:} \\ 5 children died due to wall collapse at two different places in UP's Kaushambi\end{tabular}                                                                                                                                                                                                                                                                                                                                                                                                                                                                                                                                                                                                                                                                                                                                                                                                                                                                                                          \\ \hline
\end{tabular}
\end{adjustbox}
\caption{Hindi example of headlines generated by various baseline models fine-tuned on Mukhyansh}
\label{table:Hindi Model Genrated Outputs}
\end{table*}

\begin{table*}
\begin{adjustbox}{width=\textwidth, center}
\begin{tabular}{|l|l|}
\hline
\textbf{URL}             & \url{https://telangana.suryaa.com/telangana-updates-20874-.html}                                                                                                                                                                                                                                                                                                                                                                                                                                                                                                                                                                                                                                                                                                                                                                                                                                  \\ \hline
\textbf{Article}         & \begin{tabular}[c]{@{}l@{}}\textbf{Transliteration:} \\ bījepī neta baddaṃ bāl‌rèḍḍi maraṇaṃ tīrani loṭani asèṃbli spīkar‌ pocāraṃ śrīnivāsarèḍḍi annāru. baddaṃ \\ bāl‌rèḍḍi pārthivadehānni saṃdarśiṃci nivāl̤ularpiṃcāru. anaṃtaraṃ māṭlāḍutū prajala maniṣigā bāl‌rèḍḍi \\ gurtiṃpu tèccukunnārannāru. haidarābād‌ prajalato bāl‌rèḍḍiki avinābhāva saṃbaṃdhaṃ uṃdannāru. \\ bāl‌rèḍḍi kuṭuṃba sabhyulaku tana pragāḍha sānubhūti annāru.\\ \textbf{Translation:} \\ Assembly Speaker Pocharam Srinivas Reddy termed the death of BJP leader Baddam Bal Reddy as an \\ irreparable loss. He visited the mortal remains of Baddam Bal Reddy and paid homage to him. Speaking \\ after the meeting, he said that Bal Reddy has gained recognition as a people's man. Bal Reddy has a close \\ relationship with the people of Hyderabad. My deepest condolences to the family members of Bal Reddy.\end{tabular} \\ \hline
\textbf{Actual Headline} & \begin{tabular}[c]{@{}l@{}}\textbf{Transliteration:} \\bāl‌rèḍḍi maraṇaṃ tīrani loṭu\\ \textbf{Translation:} \\Bal Reddy's death is an irreparable loss.\end{tabular}                                                                                                                                                                                                                                                                                                                                                                                                                                                                                                                                                                                                                                                                                                                                             \\ \hline
\textbf{GRU + FastText}  & \begin{tabular}[c]{@{}l@{}}\textbf{Transliteration:} \\mṛti tīrani loṭu pocāraṃ\\ \textbf{Translation:} \\Death is an irreparable loss: Pocharam\end{tabular}                                                                                                                                                                                                                                                                                                                                                                                                                                                                                                                                                                                                                                                                                                                                                     \\ \hline
\textbf{LSTM + FastText} & \begin{tabular}[c]{@{}l@{}}\textbf{Transliteration:} \\bījepī neta mṛti tīrani loṭu spīkar pocāraṃṃ\\ \textbf{Translation:} \\BJP leader's death is an irreparable loss: Speaker Pocharam\end{tabular}                                                                                                                                                                                                                                                                                                                                                                                                                                                                                                                                                                                                                                                                                                            \\ \hline
\textbf{GRU + BPEmb}     & \begin{tabular}[c]{@{}l@{}}\textbf{Transliteration:}\\ bījepī neta baddaṃ bālrèḍḍi mṛti tīraniloṭu spīkar pocāraṃ śrīnivāsarèḍḍiki saṃtāularpiṃcina asèṃbli spīkar\\ \textbf{Translation:} \\ BJP leader Baddam Bal Reddy's death is a sad loss: Assembly Speaker pays condolences to Speaker \\Pocharam Srinivasa Reddy\end{tabular}                                                                                                                                                                                                                                                                                                                                                                                                                                                                                                                                                                               \\ \hline
\textbf{mT5-small}       & \begin{tabular}[c]{@{}l@{}}\textbf{Transliteration:} \\baddaṃ bāl rèḍḍi maraṇaṃ tīrani loṭu pocāraṃ\\ \textbf{Translation:} \\Baddam Bal Reddy's death is an irreparable loss: Pocharam\end{tabular}                                                                                                                                                                                                                                                                                                                                                                                                                                                                                                                                                                                                                                                                                                              \\ \hline
\textbf{SSIB}            & \begin{tabular}[c]{@{}l@{}}\textbf{Transliteration:} \\baddaṃ bāl rèḍḍi maraṇaṃ tīrani loṭu spīkar pocāraṃ\\ \textbf{Translation:} \\Baddam Bal Reddy's death is an irreparable loss: Speaker Pocharam\end{tabular}                                                                                                                                                                                                                                                                                                                                                                                                                                                                                                                                                                                                                                                                                               \\ \hline
\end{tabular}
\end{adjustbox}
\caption{Telugu example of headlines generated by various baseline models fine-tuned on Mukhyansh}
\label{table:Telugu Model Generated Outputs}
\end{table*}

\end{document}